\documentclass[conference]{IEEEtran}

\IEEEoverridecommandlockouts
\usepackage{amsmath,amssymb,amsfonts}
\usepackage{graphicx}
 \usepackage{textcomp}
\usepackage{xcolor}
 \usepackage[ruled,vlined]{algorithm2e}
  \usepackage{bm}
 \usepackage{caption} 

\def\BibTeX{{\rm B\kern-.05em{\sc i\kern-.025em b}\kern-.08em
 T\kern-.1667em\lower.7ex\hbox{E}\kern-.125emX}}

  \makeatletter
\newcommand*{\rom}[1]{\expandafter\@slowromancap\romannumeral #1@}
\makeatother

\usepackage{xcolor} 
 
    \usepackage[left= 2.45 cm, right= 2.45 cm, top=1.778 cm, bottom = 2.65 cm]{geometry}
\usepackage{cite}
\begin{document}
\bstctlcite{IEEEexample:BSTcontrol}
\title{When Critics Disagree: Adaptive Reward Poisoning Attacks in RIS-Aided Wireless Control System}

\author{\IEEEauthorblockN{Deemah H. Tashman  and Soumaya Cherkaoui}
\IEEEauthorblockA{  \small Department of Computer  and Software Engineering, Polytechnique Montreal, Montreal, Canada\\  
Email:     \{deemah.tashman, soumaya.cherkaoui\}@polymtl.ca }}

\maketitle
\begin{abstract}
Reward-poisoning attacks present a significant risk to learning-based wireless control systems.  Given this, we propose a Disagreement-Guided Reward Poisoning (DGRP) adaptive  attack on a Soft Actor-Critic (SAC) agent. In a Cognitive Radio Network (CRN) environment assisted by Reconfigurable Intelligent Surfaces (RIS), the SAC agent is tasked with maximizing the  long-term   secondary users' (SUs) rate   by simultaneously optimizing the transmission power of the SU transmitter and the RIS phase shifts.    DGRP corrupts rewards, particularly when the SAC dual critics exhibit substantial disagreement—especially in high-leverage, high-uncertainty states—resulting in distorted value estimations and guiding the policy towards suboptimal actions.   Our findings demonstrate that DGRP substantially diminishes the performance improvements typically provided by RIS and degrades transmission quality.   We further investigate key attack parameters and determine their impact on learning.   In comparison to   periodic-timing  and exploration-triggered baselines, DGRP consistently causes greater damage, highlighting the necessity of considering disagreement-aware threats when evaluating the robustness of Deep Reinforcement Learning (DRL) in RIS-assisted networks.
\end{abstract}
\section{Introduction}

\par\IEEEPARstart{T}{o} realize the promised gains of 6G, the challenge of spectrum underutilization must be addressed to accommodate the growth in devices and connections.   Cognitive Radio Networks (CRNs) offer a viable solution, allowing unlicensed Secondary Users (SUs) to opportunistically utilize frequency bands designated for Primary Users (PUs) via underlay, overlay, or interweave modes, in accordance with regulatory constraints \cite{9237455,10368012,9838746,9500621,10278964}. When  SUs  face obstacles and unreliable connections,  conventional relays can provide assistance \cite{9926102}; yet, they lack flexibility and need substantial power, presenting difficulties for devices with constrained energy resources \cite{10361836,10466378,moudoud2023green}.   This has resulted in increased interest in passive Reconfigurable Intelligent Surfaces (RISs), which are passive reflective elements, in which a controller adjusts their phases, ideally using channel information to reshape propagation efficiently \cite{10319408}.   RISs may improve targeted signals while diminishing interference at undesired receivers by phase alignment \cite{10188924}.  In practice, however, the phase configuration must be updated for large arrays under partial, noisy, and time-varying observations with tight latency and overhead. This motivates learning-based methods that map lightweight measurements to near-optimal phase settings in real time \cite{10921906}.


Deep Reinforcement Learning (DRL) has facilitated substantial advancements in control and optimization  \cite{11370849,10078092,9685236,9839316,10592377,9999295,10243611,9500285,10437455,10433640} for CRNs and RIS-assisted systems.    Nonetheless, with these advantages, the security of DRL—especially its vulnerability to reward poisoning—has emerged as a critical concern that can hinder training and reduce deployment effectiveness \cite{xu2022efficient,11371394}.    In CRNs, even slight manipulation of rewards can result in erroneous decisions on spectrum access, leading to detrimental interference  and a decline in reliability and user trust.    Most recent studies on reward poisoning use basic timing protocols or initial exploration triggers that do not adjust to the agent's learning condition or prioritize the most significant updates.  For instance, in \cite{bouhaddi2023multi}, a white-box reward-poisoning attack that injects bounded perturbations into the agent’s observed rewards at  attacker-selected (targeted) timesteps/states  was proposed. In \cite{xu2023black}, the authors proposed a black-box targeted reward-poisoning attack by offsetting the rewards to steer the learner toward an attacker-specified policy under per-step and total budgets, intervening only when the agent’s chosen action deviates from the target policy. Moreover, the authors  proposed a white-box, exploration-guided reward-poisoning attack that  alters the reward  at timesteps flagged by high exploration in \cite{cai2023reward}. Furthermore, in \cite{xu2022efficient}, uniform random and policy-guided variants that add bounded reward offsets at selected steps under strict per-step and total-step budgets were proposed. In addition to the limited poisoning attacks studies for DRL, the existing ones are not state-adaptive—they neither read the learner’s uncertainty nor focus perturbations on high-leverage, high-uncertainty states, instead relying on fixed timing or coarse exploration heuristics. Moreover, prior studies did not consider  realistic wireless communication environments; they evaluated attacks in simplified testbeds rather than practical wireless settings.

To the best of the authors' knowledge, no previous work has examined a state-adaptive reward poisoning attack on a DRL agent within a realistic wireless scenario.   Given this, the main contributions of this work are given as follows:
\begin{itemize} 
\item  In a CRN environment assisted by a passive RIS, in which a Soft Actor-Critic (SAC) agent is  tasked with maximizing long-term SUs' rate, we propose a Disagreement-Guided Reward Poisoning (DGRP) attack that is state-adaptive, targeting the SAC agent to hinder the learning process and reduce the SUs' data rate.

\item  The proposed DGRP attack is assumed to corrupt  rewards precisely when the twin critics exhibit large disagreement (high-leverage, high-uncertainty states), distorting value estimation and steering the policy toward suboptimal actions.  

\item We study the impact of our attack on the RIS efficiency, and we compare it with  several state-of-the-art baselines, such as   Periodic Timing, and Exploration-Triggered attacks, showing DGRP induces the largest performance degradation.
\end{itemize}

The remaining sections of the paper are presented as follows: the DRL agent test environment is illustrated in section \rom{2}. The SUs' rate maximization problem and the DRL solution are provided in section \rom{3}. Section \rom{4} contains the proposed DGRP attack.   Finally, the results are presented in section \rom{5}, and the conclusions are included in section \rom{6}.

 \section{Deep Reinforcement Learning Test Environment}
\begin{figure}[hb]
  \centering
  \includegraphics[width=1.0\linewidth]{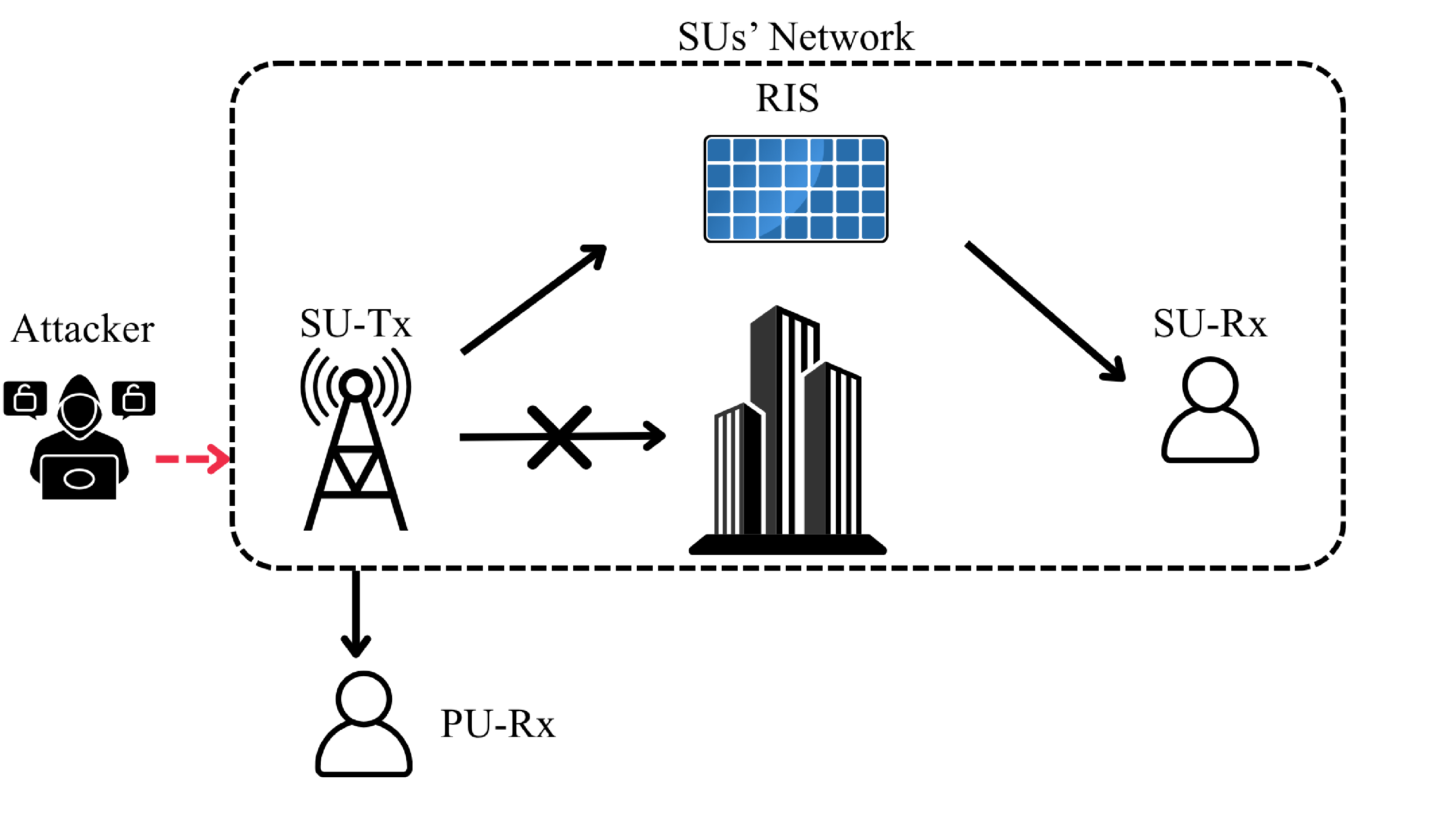}
  \caption{The system model.}
  \label{sys1}
\end{figure}

We employ a CRN scenario to evaluate the effects of the reward poisoning attack on the SAC agent. Fig.  \ref{sys1} illustrates an underlay cognitive radio configuration, whereby a secondary transmitter (SU-Tx)   transmits signals to a secondary user receiver (SU-Rx).   Due to the presumed extensive blockage of direct SUs link \cite{bae2024overview}, communication is facilitated by an RIS affixed to a building, which comprises passive reflecting elements ($R$) managed by a lightweight controller to direct waves toward the designated SU receiver (SU-Rx).   The SUs operate in underlay mode, allowing the SU-Tx to transmit data concurrently with the PUs, provided that the transmission power maintains interference at the primary receiver (PU-Rx) below a specified threshold \cite{11059714}. The channel gain from the SU transmitter to the RIS is characterized by Rayleigh fading and denoted by $\mathbf{H_{1}} \in \mathbb{C}^{1 \times R}$.  Likewise, the channel vector between the RIS and the SU-Rx is also characterized by Rayleigh fading and  given as $\mathbf{H_{2}} \in \mathbb{C}^{R \times 1}$.  Therefore, the signal received at SU-Rx is expressed as
 \begin{IEEEeqnarray}{lCr} \label{ys} 
y_s= \sqrt{P_s} \left(\mathbf{H_{1} \Phi H_{2}}\right)   x +n_s,
\end{IEEEeqnarray}
\noindent where $x$ is the message transmitted by the SU-Tx, $P_s$ is the transmission power of the SU-Tx, $\mathbf{\Phi}=\operatorname{diag}[\phi_1,\phi_2,\cdots,\phi_R]$   signifies the diagonal phase shift matrix of the RIS, and  $n_s$ represents the Additive White Gaussian Noise (AWGN) at the SU-Rx. We assume an ideal passive RIS with unit-modulus elements (no amplification) \cite{10561503}; hence, the entries of $\mathbf{\Phi}$ are given as $\phi_r=\beta_r e^{j\rho_r}$,   such that $\beta_r=1$ and $\rho_r\in[0,2\pi)$.
Given (\ref{ys}), the Signal-to-Noise Ratio (SNR) at SU-Rx is expressed as
\begin{IEEEeqnarray}{lCr}  \label{snrs}
\lambda_s=\frac{ P_s \left| \mathbf{H_{1} \Phi H_{2}  }  \right|^2}{  N_0 },
\end{IEEEeqnarray}
\noindent where  $N_0$ represents the AWGN variance at the SU-Rx.   The SUs data rate is utilized to assess the system's performance as
\begin{IEEEeqnarray}{lCr}  \label{rate}
 R_s= \log_2 \left(1+\lambda_s\right).
\end{IEEEeqnarray}

The SU-Tx must restrict its transmit power to ensure that the interference experienced at the PU-Rx remains within the established threshold. Hence, the constraint is expressed as \cite{9562609,9348134,9408651,9612017}
\begin{IEEEeqnarray}{lCr}  \label{const}
P_s \leq \min \{P_{m},\frac{I}{g_{p}}\}, 
\end{IEEEeqnarray}
\noindent where \( P_{m} \) denotes the maximum power permissible for the SU-Tx, \( I \) represents the maximum interference tolerable by the PU-Rx, and \( g_{p} \) signifies the channel power gain for the SU-Tx — PU-Rx link \( (h_{p}) \), which is characterized by Rayleigh fading. 

\section{SUs' Rate Maximization Problem \& DRL-based Solution}
\subsection{Problem Formulation:}
Our goal is to jointly tune the SU transmit power and the RIS phase shifts to maximize the SUs’ rate, subject to constraints that ensure the PUs communication is not degraded. Accordingly, we formulate the following optimization problem
\begin{align} \label{opti-prob}
   \mathcal{P}: \;\;  & \underset{\mathbf{\Phi},P_s}{\text{max}}
    & &  R_s \\
    & \text{s.t.}  
    && \label{firstrate1}P_s \leq \min \{P_{m},\frac{I}{g_{p}}\}, &  \\ &&& \label{secondcon}  |\rho_r|\in[0,2\pi),  \forall r=1,2,\cdots,R. & 
  \end{align}
 
\subsection{Soft-Actor Critic DRL Solution:}
We designate the SU-Tx as a DRL agent tasked with resolving the optimization problem in (\ref{opti-prob}) by maximizing the rate of the SUs through the joint control of its transmit power and the RIS phase shifts.   The dynamics conform to the Markov property, where each state depends exclusively on its prior state; thus, we employ a model-free Markov decision process (MDP) framework \cite{10182973,10622458,9318243,9729992,9524882,10646359,10794361}. At time $t$, the \textit{state space}  combines the interference threshold at the PU-Rx, the pertinent channels, and the most recent control applied, i.e.
$ \mathcal{S}_{t}=\{ I^{t}, \mathbf{H_1^{t}}, \mathbf{H_2^{t}}, h_{p}^{t}, P_{s}^{(t-1)}, \mathbf{\Phi^{(t-1)}} \} $. At each state, the agent selects an action from the \textit{action space} at time $t$, which is composed of the new transmit power and RIS phases, $\mathcal{A}_{t}=\{ P_{s}^{t},  \mathbf{\Phi}^{t} \}.$
After executing $A_{t}$ in $S_{t}$, the agent obtains a  \textit{reward}   equal to the SUs' rate, 
$\mathcal{R}_{t}=R_s$

Two primary considerations justify the suitability of SAC for our context.  First, the problem involves continuous states and actions, and SAC is designed to perform effectively in continuous control scenarios.   Second, SAC has regularly exceeded prior DRL benchmarks in comparable wireless and communication operations \cite{10283517}.  SAC employs three neural networks: a stochastic actor (policy) that translates states into a distribution of actions and two critic Q-networks that evaluate the expected return for state-action pairs.   The twin-critic design reduces overestimation bias, while the stochastic actor enables efficient exploration in continuous action spaces. Given this, the Q-networks in the SAC technique are jointly trained as \cite{10283517}
   \begin{IEEEeqnarray}{lcr} 
  U=r+\gamma\min_{i=1,2} Q_{\theta_{i}'}\left(s',a'\right)-e_s \log \left(a'|s'\right) , 
\end{IEEEeqnarray} 
   \begin{IEEEeqnarray}{lcr} 
   Z(\theta_i)=\frac{1}{M}\|U-Q_{\theta_{i}} (s,a)\|_2^2, 
\end{IEEEeqnarray} 
   \begin{IEEEeqnarray}{lcr} 
    \theta_i  \leftarrow \theta_i- l \nabla_{\theta_{i}} Z(\theta_i) , 
\end{IEEEeqnarray}

\noindent where $\theta_i$ denote the parameters of the $i$-th Q-network, $M$ is the mini-batch size sampled from the replay buffer $(\mathcal{D})$, and $l$ the learning rate. $Z(\theta_i)$ denotes  the critic loss, and $\nabla_{\theta_i} Z(\theta_i)$ represents its gradient with respect to $\theta_i$. The entropy coefficient ($e_s$) dictates the extent of exploration by SAC; a greater $e_s$ value results in action distributions with higher entropy, indicating increased exploration.   Deterministic-policy approaches often incorporate external noise for exploration; however, SAC employs entropy regularization with automatic entropy adjustment to modify $e_s$ in real-time, contingent upon the current policy's entropy relative to a predetermined target value.   This process balances exploration and exploitation without necessitating the manual introduction of noise.  This is particularly advantageous in our scenario (combined transmit power and RIS phase optimization), where extensive yet efficient exploration accelerates convergence and enhances final performance.

The policy network employs state vectors as input and generates action vectors as output.  The loss function for this network can be expressed as
   \begin{IEEEeqnarray}{lcr} 
  Z(\zeta) = \frac{1}{M} \sum_{j=1}^{M} \left( e_s \log \pi_{\zeta}(a_{j}|s_{j}) - \min_{i=1,2} Q_{\theta_{i}}(s_{j}, a_{j}) \right), \nonumber \\ 
\end{IEEEeqnarray}
\noindent where $\pi_{\zeta}(a_j|s_j)$ denotes the stochastic policy—parameterized by $\zeta$—that returns the likelihood of selecting action  $a_j$  in state  $s_j$. The policy gradient ($\nabla_{\zeta} Z(\zeta)$) is then estimated via the stochastic policy-gradient method and used to update  $\zeta$  by gradient ascent, i.e., moving  $\zeta$  in the direction of  $\nabla_{\zeta} Z(\zeta)$  as 
   \begin{IEEEeqnarray}{lcr} 
   \zeta \leftarrow \zeta + l \nabla_\zeta Z(\zeta).
\end{IEEEeqnarray}
\noindent It is important to note that we train in a non-episodic (continuing) environment, as the task lacks a natural terminal state, and as a result, we optimize the long-term average reward to encourage consistent learning and performance over time.

{\color{black} \section{Disagreement-Guided Reward Poisoning in Deep Reinforcement Learning}
We propose \textit{Disagreement-Guided Reward Poisoning} (DGRP)—a stealthy, training-time reward-poisoning attack against SAC in the above CRN setting.   In this attack, the attacker does not alter observations or actions; rather, they only make minor, sparse modifications to the reward signal during the most sensitive period of learning.   The primary signal is "critic disagreement",  which is the discrepancy between SAC's two Q-networks on the current $(s_t,a_t)$.   The attacker declares step $t$ eligible when the gap of the disagreement exceeds a specific percentile threshold by employing a rolling window that stores the most recent $w$  critic-gap value. Conditional on eligibility, the attacker fires stochastically with probability $p$ (the budget) and then subtracts a bounded amount of $\delta$ from the reward before the agent’s reward normalization.      The attack model is illustrated in Fig. \ref{sys2}. 

Let $r_t^{\text{true}}$ denotes the environmental (clean) reward at step $t$. After the attacker injects small, targeted corruptions to the reward, the reward value that is actually utilized for learning is expressed as
\begin{equation}
  r_t^{\text{train}} \;=\; r_t^{\text{true}} \;-\; \delta\,u_t, \qquad \delta>0,
\end{equation}
where $u_t\in\{0,1\}$ indicates whether an attack is applied and $\delta$ is a bounded magnitude.

   \begin{figure} 
  \centering
  \includegraphics[width=1.0\linewidth]{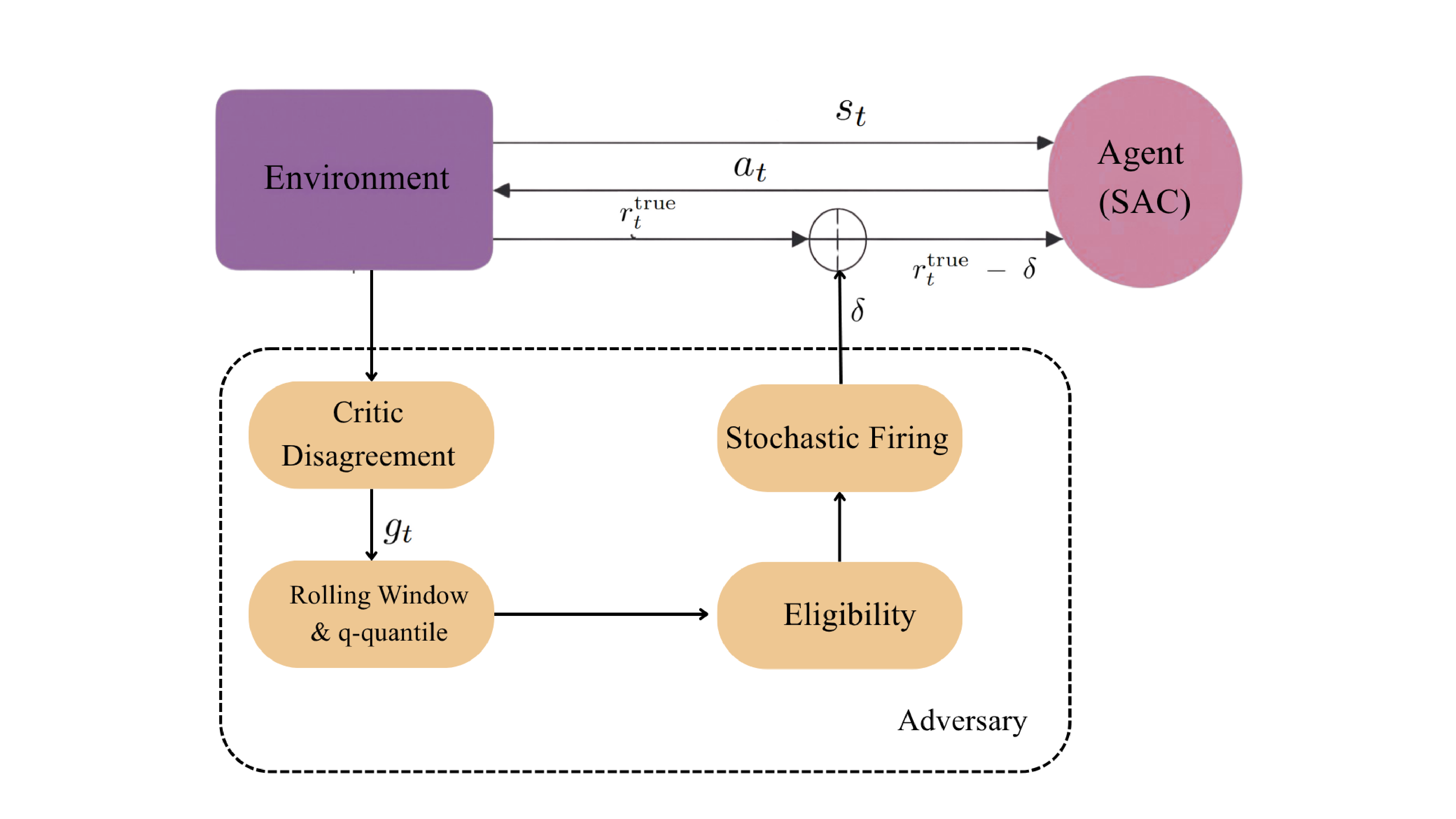}
  \caption{The attack model.}
  \label{sys2}
\end{figure}

\newcommand{\train}{\mathrm{train}}
\newcommand{\true}{\mathrm{true}}
\newcommand{\Pct}{\mathrm{Pct}}
\newcommand{\EMA}{\mathrm{EMA}}
\newcommand{\DGRP}{\textsc{DGRP}\xspace}

To mount the attack, we assume the adversary has (gray/white-box) access to the outputs of the agent’s twin critics  ($Q_1,Q_2$) on the current state–action pair \cite{xu2022efficient}.  They do not require any information regarding the environment, except for the reward that was observed. The disagreement signal is defined as
\begin{equation}
  g_t \;=\; \bigl\lVert Q_1(s_t,a_t) - Q_2(s_t,a_t) \bigr\rVert.
\end{equation}
\noindent We employ the critic disagreement notion since  a large gap indicates substantial epistemic uncertainty and gradient sensitivity in SAC.   Altering rewards throughout these periods has a significant impact while remaining sparse and hard to detect.

We keep a fixed-length rolling buffer ($\mathcal{G}_t$) of the last $w$ disagreement values to adapt to the agent's learning dynamics and the current environment.   At each step, we set the threshold to the empirical $q$-quantile of this buffer,
$[
\tau_t = \mathrm{Quantile}_q(\mathcal{G}_t)
]$, and declare the step \emph{eligible} if ($g_t >\tau_t$). This rule is self-tuning and scale-free: when disagreement becomes larger or more variable, $\tau_t$ increases; when it stabilizes, $\tau_t$ decreases.  
 
After an eligibility check (i.e., the step meets the gap-based criterion), the attacker  fires with probability  $p$  on that eligible step. Indeed,  $p$ controls how often the attack is applied among the eligible moments: only a  portion  of eligible steps are actually poisoned. That is,  the overall poisoning rate equals “fraction of eligible steps $\times$ $p$.” This two-stage design separates  \textit{when}  an attack can happen (governed by critic disagreement) from  \textit{how frequently}  it does (governed by $p$), avoiding a fixed pattern and making the attack harder to detect or counter. The attack procedure is summarized in Algorithm~\ref{alg:DGRP}.


 \begin{algorithm}[h]
\caption{SAC Training with Disagreement-Guided Reward Poisoning (\DGRP)}
\KwIn{SAC hyperparameters; replay buffer $\mathcal{D}$; \DGRP params: window $w$, quantile $q$, budget $p$, magnitude $\delta$, warm-up $T_{\text{warm}}$}
\KwOut{Trained policy for each random seed}

\BlankLine
\For{each random seed}{
  Initialize actor, twin critics $(Q_1,Q_2)$, temperature; buffer $\mathcal{D}$; rolling window $\mathcal{G}\!\leftarrow\!\emptyset$\;

  \Repeat{convergence}{
    Observe $s_t$; sample $a_t$ from policy; step env to get $s_{t+1}$ and $r_t^{\true}$\;

    \tcp{Stage 1: compute eligibility}
    Compute $g_t \leftarrow \lVert Q_1(s_t,a_t)-Q_2(s_t,a_t)\rVert$; push $g_t$ into $\mathcal{G}$ (cap size $w$)\;
    Define $\texttt{eligible} \leftarrow \mathbf{1}\{\,t \ge T_{\text{warm}} \land |\mathcal{G}|=w \land g_t > \mathrm{Quantile}_q(\mathcal{G})\,\}$\;

    \tcp{Stage 2: stochastic firing}
    {On eligible steps, attack with probability $p$.} Set $\texttt{fire} \leftarrow \texttt{eligible} \cdot \mathbf{1}\{\text{attack}\}$\;

    \tcp{Apply corruption iff $\texttt{fire}=1$}
    Set $r_t^{\train} \leftarrow r_t^{\true} - \delta \cdot \texttt{fire}$\;

    Store $(s_t,a_t,r_t^{\train},s_{t+1},\texttt{done})$ in $\mathcal{D}$\;
    Update critics, policy, and temperature (SAC);\quad $s_t \leftarrow s_{t+1}$\;
  }
}
\label{alg:DGRP}
\end{algorithm}

 \section{Simulation Results}
In this section, we present the simulation results and analyses for the considered system model. We follow common DRL benchmarking practice by running each experiment over ten independent random seeds, consistent with the protocol in \cite{10283517}. The parameters are set to the following values, unless otherwise specified in the captions: $w=200$, $T_{\text{warm}}=50$, $q=0.75$,   $\gamma=1.0$, $l=10^{-3}$, $e_s=0.2$, $\mathcal{D}=2\!\times\!10^{4}$,  $M=16$,     $P_{m}=1$ dB, $I=10$ dB,  and  $N_0=10^{-2}$.

Fig.  \ref{fig1} illustrates the average rate of SUs across various numbers of RIS elements ($R$) and attack magnitudes ($\delta$). Augmenting $\delta$ consistently reduces the learning curve.  This is because the agent receives a more adverse and noisy  reward signal, rendering advantageous actions appear less significant and prompting cautious policy modifications.      The effect of the proposed attack  is most evident when comparing the "No Attack" curves to the attacked reward curves.   In clean training, augmenting the RIS from $R=6$ to $R=16$ results in a substantial rate enhancement due to the agent's ability to exploit additional spatial degrees of freedom.   In the presence of reward poisoning, the benefits diminish: despite an improved environment, negative feedback complicates the agent's ability to consistently associate high returns with appropriate transmission power  and phase selections.   In summary, reward attacks alter the objective by systematically reducing the reward by $\delta$.  Together, these factors impede the policy's ability to leverage larger $R$ and ultimately constrain the maximum data rate.

\begin{figure}     
  \centering
  \includegraphics[width=1.0\linewidth]{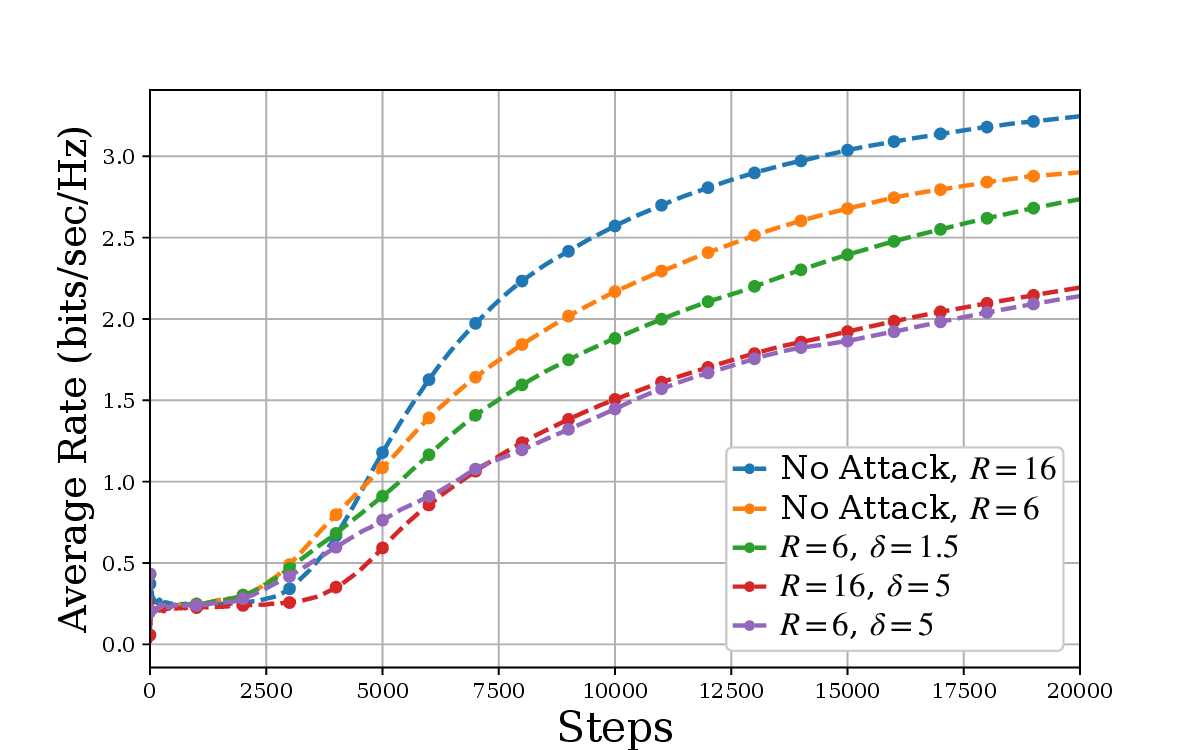}
  \caption{Average reward of SUs' for different values of $R$ and $\delta$.}
  \label{fig1}
\end{figure}

In Fig. \ref{fig2}, we compare our proposed approach to two cases; No Attack and Periodic Timing Attack.     For Periodic-timing attack, the attacker disrupts the reward at a fixed schedule (every two steps, with the phase and power varying randomly each episode), resulting in consistent, evenly spaced changes.   Corruptions in Periodic-timing occur at predetermined intervals, resulting in a persistent distortion of return estimates.  This indicates that the agent receives erroneous feedback at predictable points during the training.  Our DGRP  attack is more detrimental due to its state-adaptive nature; it activates when the two critics exhibit significant disagreement (high uncertainty).  Corrupting rewards in these high-impact states significantly alters the objective and drives policy towards suboptimal actions.  We can conclude from this figure that DGRP must be explicitly included when evaluating reward-poisoning resilience in DRL systems utilizing RIS control, since basic timing-based benchmarks do not adequately reflect the worst-case scenarios.

 \begin{figure}    [b] 
  \centering
  \includegraphics[width=1.0\linewidth]{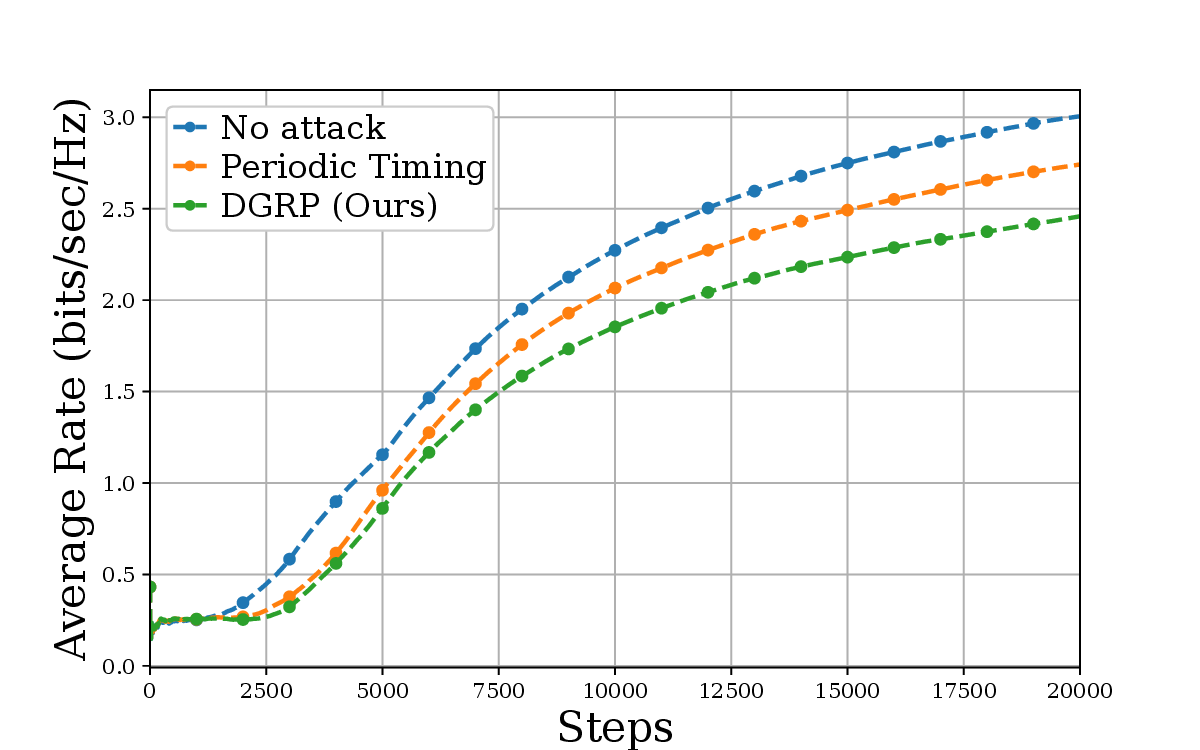}
  \caption{A comparison between No Attack,  Periodic Timing, and DGRP (our proposed attack) in terms of the average SUs' rate. $\delta=1.5$, $R=6$, and  $p=0.5$.}
  \label{fig2}
\end{figure} 

Fig. \ref{fig3} illustrates the average rate of SUs in relation to $p$. For the proposed DGRP attack, as  $p$ increases, a greater number of time steps get compromised, resulting in a more pronounced negative bias in return estimates and policy adjustments.  This results in a consistent decline in the average rate of the SUs. We additionally contrast our attack with the Exploration-triggered attack from prior research \cite{cai2023reward}.  This attack predominantly occurs during high-entropy exploration, primarily in the initial training phase, and not exclusively when the value is crucial.   As exploration diminishes, the opportunities for effective firing decrease, resulting in diminished impact.   DGRP, conversely, persists in targeting powerful situations during training, rendering it significantly more detrimental than the exploration-based attack.

 \begin{figure}     
  \centering
  \includegraphics[width=1.0\linewidth]{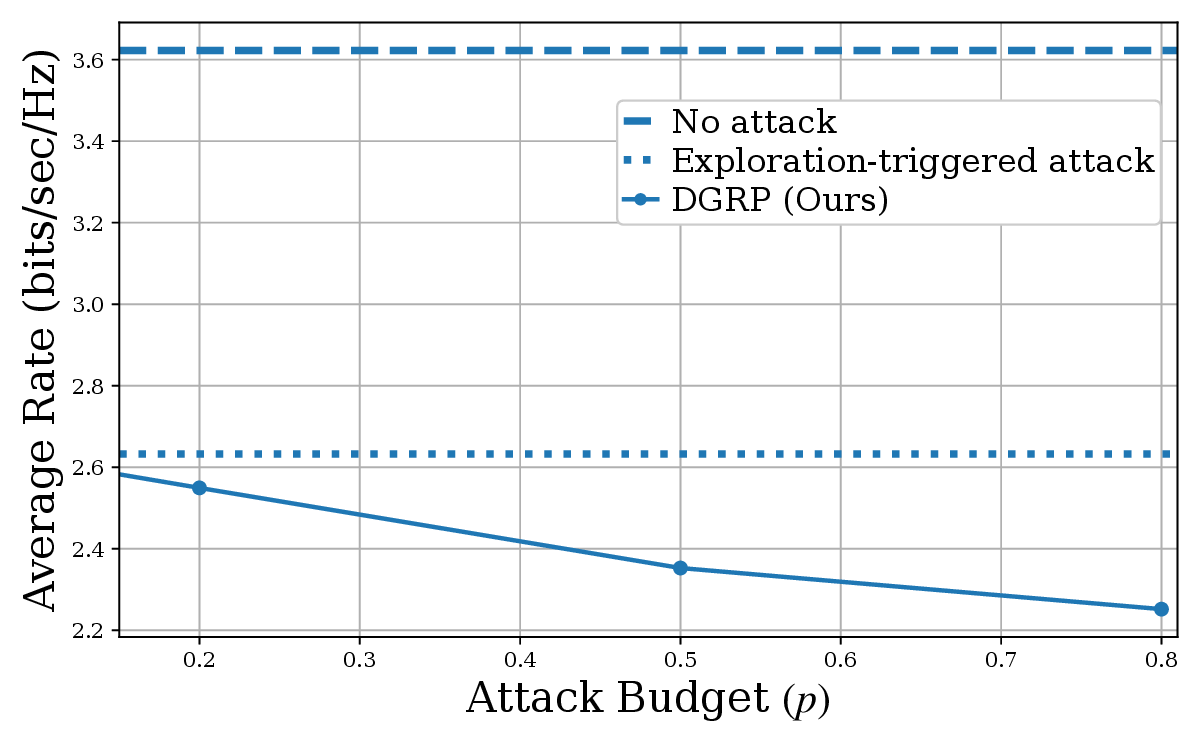}
  \caption{A comparison between No Attack,   Exploration-triggered attack, and DGRP (our proposed attack) in terms of the average SUs' rate. $\delta=1.5$, and $R=6$.}
  \label{fig3}
\end{figure} 

\section{Conclusions}
In this paper, we proposed a  Disagreement-Guided Reward Poisoning (DGRP)  attack on a Soft Actor–Critic (SAC) agent within a cognitive radio network utilizing Reconfigurable Intelligent Surfaces (RIS).   The attacker compromises rewards when the twin critics exhibit the greatest disagreement, indicating states where the  $Q$-value disparity is maximal.  This addresses the most sensitive periods of learning.   Under DGRP, the RIS gains fail to appear as the agent receives persistently misleading feedback.   DGRP causes larger and more persistent degradation in long-term return compared to    Periodic-timing  baseline, highlighting that timing-only benchmarks may fail to accurately assess worst-case susceptibility.   We also characterize DGRP by its  magnitude and  budget, indicating that an increase in their values causes performance degradation and diminishes the average rate of SUs.   Ultimately,  compared to a baseline from earlier work where attack  was triggered by exploration, DGRP is practical and more damaging, as it focuses on critical stages throughout the training process rather than solely during the first training phases.


\begin{thebibliography}{10}
\providecommand{\url}[1]{#1}
\csname url@samestyle\endcsname
\providecommand{\newblock}{\relax}
\providecommand{\bibinfo}[2]{#2}
\providecommand{\BIBentrySTDinterwordspacing}{\spaceskip=0pt\relax}
\providecommand{\BIBentryALTinterwordstretchfactor}{4}
\providecommand{\BIBentryALTinterwordspacing}{\spaceskip=\fontdimen2\font plus
\BIBentryALTinterwordstretchfactor\fontdimen3\font minus \fontdimen4\font\relax}
\providecommand{\BIBforeignlanguage}[2]{{%
\expandafter\ifx\csname l@#1\endcsname\relax
\typeout{** WARNING: IEEEtran.bst: No hyphenation pattern has been}%
\typeout{** loaded for the language `#1'. Using the pattern for}%
\typeout{** the default language instead.}%
\else
\language=\csname l@#1\endcsname
\fi
#2}}
\providecommand{\BIBdecl}{\relax}
\BIBdecl

\bibitem{9237455}
D.~H. Tashman \emph{et~al.}, ``{An Overview and Future Directions on Physical-Layer Security for Cognitive Radio Networks},'' \emph{IEEE Network}, vol.~35, no.~3, pp. 205--211, 2021.

\bibitem{10368012}
N.~A. Khalek \emph{et~al.}, ``{Advances in Machine Learning-Driven Cognitive Radio for Wireless Networks: A Survey},'' \emph{IEEE Communications Surveys \& Tutorials}, vol.~26, no.~2, pp. 1201--1237, 2024.

\bibitem{9838746}
D.~H. Tashman \emph{et~al.}, ``{Towards Improving the Security of Cognitive Radio Networks-Based Energy Harvesting},'' in \emph{ICC 2022 - IEEE International Conference on Communications}, 2022, pp. 3436--3441.

\bibitem{9500621}
D.~H. Tashman \emph{et~al.}, ``{Secrecy Analysis for Energy Harvesting-Enabled Cognitive Radio Networks in Cascaded Fading Channels},'' in \emph{ICC 2021 - IEEE International Conference on Communications}, 2021, pp. 1--6.

\bibitem{10278964}
D.~H. Tashman \emph{et~al.}, ``{Securing Cognitive Radio Networks via Relay and Jammer-Based Energy Harvesting on Cascaded Channels},'' in \emph{ICC 2023 - IEEE International Conference on Communications}, 2023, pp. 3246--3251.

\bibitem{9926102}
D.~H. Tashman \emph{et~al.}, ``{Overlay Cognitive Radio Networks Enabled Energy Harvesting With Random AF Relays},'' \emph{IEEE Access}, vol.~10, pp. 113\,035--113\,045, 2022.

\bibitem{10361836}
H.~Zhou \emph{et~al.}, ``{A Survey on Model-Based, Heuristic, and Machine Learning Optimization Approaches in RIS-Aided Wireless Networks},'' \emph{IEEE Communications Surveys \& Tutorials}, vol.~26, no.~2, pp. 781--823, 2024.

\bibitem{10466378}
D.~H. Tashman \emph{et~al.}, ``{Maximizing Reliability in Overlay Radio Networks With Time Switching and Power Splitting Energy Harvesting},'' \emph{IEEE Transactions on Cognitive Communications and Networking}, vol.~10, no.~4, pp. 1307--1316, 2024.

\bibitem{moudoud2023green}
H.~Moudoud \emph{et~al.}, ``{Green machine learning for Internet-of-Things: Current solutions and future challenges},'' in \emph{Green Machine Learning Protocols for Future Communication Networks}.\hskip 1em plus 0.5em minus 0.4em\relax CRC Press, 2023, pp. 161--175.

\bibitem{10319408}
L.~Wang \emph{et~al.}, ``{Hybrid Hierarchical DRL Enabled Resource Allocation for Secure Transmission in Multi-IRS-Assisted Sensing-Enhanced Spectrum Sharing Networks},'' \emph{IEEE Transactions on Wireless Communications}, vol.~23, no.~6, pp. 6330--6346, 2024.

\bibitem{10188924}
W.~Khalid \emph{et~al.}, ``{Reconfigurable Intelligent Surface for Physical Layer Security in 6G-IoT: Designs, Issues, and Advances},'' \emph{IEEE Internet of Things Journal}, vol.~11, no.~2, pp. 3599--3613, 2024.

\bibitem{10921906}
D.~H. Tashman \emph{et~al.}, ``{Dynamic Synergy: Leveraging RIS and Reinforcement Learning for Secure, Adaptive Underlay Cognitive Radio Networks},'' in \emph{2025 Global Information Infrastructure and Networking Symposium (GIIS)}, 2025, pp. 1--6.

\bibitem{11370849}
M.~C. Kirana \emph{et~al.}, ``{ML-Enabled Open RAN: A Comprehensive Survey of Architectures, Challenges, and Opportunities},'' \emph{IEEE Communications Surveys \& Tutorials}, pp. 1--1, 2026.

\bibitem{10078092}
A.~Filali \emph{et~al.}, ``{Communication and Computation O-RAN Resource Slicing for URLLC Services Using Deep Reinforcement Learning},'' \emph{IEEE Communications Standards Magazine}, vol.~7, no.~1, pp. 66--73, 2023.

\bibitem{9685236}
A.~Abouaomar \emph{et~al.}, ``{Mean-Field Game and Reinforcement Learning MEC Resource Provisioning for SFC},'' in \emph{2021 IEEE Global Communications Conference (GLOBECOM)}, 2021, pp. 1--6.

\bibitem{9839316}
Z.~Mlika \emph{et~al.}, ``{Deep Deterministic Policy Gradient to Minimize the Age of Information in Cellular V2X Communications},'' \emph{IEEE Trans. Intell. Transp. Syst.}, vol.~23, no.~12, pp. 23\,597--23\,612, 2022.

\bibitem{10592377}
D.~H. Tashman \emph{et~al.}, ``Securing next-generation networks against eavesdroppers: Fl-enabled drl approach,'' in \emph{2024 International Wireless Communications and Mobile Computing (IWCMC)}, 2024, pp. 1643--1648.

\bibitem{9999295}
A.~Abouaomar \emph{et~al.}, ``{Federated Deep Reinforcement Learning for Open RAN Slicing in 6G Networks},'' \emph{IEEE Communications Magazine}, vol.~61, no.~2, pp. 126--132, 2023.

\bibitem{10243611}
H.~Moudoud \emph{et~al.}, ``{Empowering Security and Trust in 5G and Beyond: A Deep Reinforcement Learning Approach},'' \emph{IEEE Open Journal of the Communications Society}, vol.~4, pp. 2410--2420, 2023.

\bibitem{9500285}
Z.~Mlika \emph{et~al.}, ``{Competitive Algorithms and Reinforcement Learning for NOMA in IoT Networks},'' in \emph{ICC 2021 - IEEE International Conference on Communications}, 2021, pp. 1--6.

\bibitem{10437455}
Y.~Tao \emph{et~al.}, ``{Digital Twin and DRL-Driven Semantic Dissemination for 6G Autonomous Driving Service},'' in \emph{GLOBECOM 2023 - 2023 IEEE Global Communications Conference}, 2023, pp. 2075--2080.

\bibitem{10433640}
A.~Filali \emph{et~al.}, ``{Open RAN Slicing for MVNOs With Deep Reinforcement Learning},'' \emph{{IEEE Internet of Things Journal}}, vol.~11, no.~10, pp. 18\,711--18\,725, 2024.

\bibitem{xu2022efficient}
Y.~Xu \emph{et~al.}, ``{Efficient reward poisoning attacks on online deep reinforcement learning},'' \emph{arXiv preprint arXiv:2205.14842}, 2022.

\bibitem{11371394}
D.~H. Tashman \emph{et~al.}, ``{Trustworthy AI-Driven Dynamic Hybrid RIS: Joint Optimization and Reward Poisoning-Resilient Control in Cognitive MISO Networks},'' \emph{IEEE Transactions on Network and Service Management}, pp. 1--1, 2026.

\bibitem{bouhaddi2023multi}
M.~Bouhaddi \emph{et~al.}, ``{Multi-Environment Training Against Reward Poisoning Attacks on Deep Reinforcement Learning.}'' in \emph{SECRYPT}, 2023, pp. 870--875.

\bibitem{xu2023black}
Y.~Xu \emph{et~al.}, ``{Black-box targeted reward poisoning attack against online deep reinforcement learning},'' \emph{arXiv preprint arXiv:2305.10681}, 2023.

\bibitem{cai2023reward}
K.~Cai \emph{et~al.}, ``{Reward poisoning attacks in deep reinforcement learning based on exploration strategies},'' \emph{Neurocomputing}, vol. 553, p. 126578, 2023.

\bibitem{bae2024overview}
J.~Bae \emph{et~al.}, ``{Overview of RIS-enabled secure transmission in 6G wireless networks},'' \emph{Digital Communications and Networks}, 2024.

\bibitem{11059714}
D.~H. Tashman \emph{et~al.}, ``{Quantum-Aided Active User Detection for Energy-Efficient CD-NOMA in Cognitive Radio Networks},'' in \emph{2025 International Wireless Communications and Mobile Computing (IWCMC)}, 2025, pp. 1661--1666.

\bibitem{10561503}
J.~Singh \emph{et~al.}, ``{Joint Hybrid Transceiver and Reflection Matrix Design for RIS-Aided mmWave MIMO Cognitive Radio Systems},'' \emph{IEEE Transactions on Cognitive Communications and Networking}, vol.~11, no.~1, pp. 391--407, 2025.

\bibitem{9562609}
N.~D. Nguyen \emph{et~al.}, ``{Secrecy Outage Probability of Reconfigurable Intelligent Surface-Aided Cooperative Underlay Cognitive Radio Network Communications},'' in \emph{2021 22nd Asia-Pacific Network Operations and Management Symposium (APNOMS)}, 2021, pp. 73--77.

\bibitem{9348134}
D.~H. Tashman \emph{et~al.}, ``{Physical-Layer Security for Cognitive Radio Networks over Cascaded Rayleigh Fading Channels},'' in \emph{GLOBECOM 2020 - 2020 IEEE Global Communications Conference}, 2020, pp. 1--6.

\bibitem{9408651}
D.~H. Tashman \emph{et~al.}, ``{Physical-Layer Security on Maximal Ratio Combining for SIMO Cognitive Radio Networks Over Cascaded $\kappa$-$\mu$ Fading Channels},'' \emph{IEEE Transactions on Cognitive Communications and Networking}, vol.~7, no.~4, pp. 1244--1252, 2021.

\bibitem{9612017}
D.~H. Tashman \emph{et~al.}, ``{On Securing Cognitive Radio Networks-Enabled SWIPT Over Cascaded $\kappa$-$\mu$ Fading Channels With Multiple Eavesdroppers},'' \emph{IEEE Transactions on Vehicular Technology}, vol.~71, no.~1, pp. 478--488, 2022.

\bibitem{10182973}
D.~H. Tashman \emph{et~al.}, ``{Performance Optimization of Energy-Harvesting Underlay Cognitive Radio Networks Using Reinforcement Learning},'' in \emph{2023 International Wireless Communications and Mobile Computing (IWCMC)}, 2023, pp. 1160--1165.

\bibitem{10622458}
D.~H. Tashman \emph{et~al.}, ``{Federated Learning-based MARL for Strengthening Physical-Layer Security in B5G Networks},'' in \emph{ICC 2024 - IEEE International Conference on Communications}, 2024, pp. 293--298.

\bibitem{9318243}
Z.~Mlika \emph{et~al.}, ``{Network Slicing with MEC and Deep Reinforcement Learning for the Internet of Vehicles},'' \emph{IEEE Network}, vol.~35, no.~3, pp. 132--138, 2021.

\bibitem{9729992}
A.~Filali \emph{et~al.}, ``{Dynamic SDN-Based Radio Access Network Slicing With Deep Reinforcement Learning for URLLC and eMBB Services},'' \emph{IEEE Trans. Network Sci. Eng.}, vol.~9, no.~4, pp. 2174--2187, 2022.

\bibitem{9524882}
A.~Abouaomar \emph{et~al.}, ``{A Deep Reinforcement Learning Approach for Service Migration in MEC-enabled Vehicular Networks},'' in \emph{2021 IEEE 46th Conference on Local Computer Networks (LCN)}, 2021, pp. 273--280.

\bibitem{10646359}
D.~H. Tashman \emph{et~al.}, ``{Optimizing Cognitive Networks: Reinforcement Learning Meets Energy Harvesting Over Cascaded Channels},'' \emph{IEEE Systems Journal}, vol.~18, no.~4, pp. 1839--1848, 2024.

\bibitem{10794361}
D.~H. Tashman \emph{et~al.}, ``{Securing Cognitive IoT Networks: Reinforcement Learning for Adaptive Physical Layer Defense},'' in \emph{2024 6th International Conference on Communications, Signal Processing, and their Applications (ICCSPA)}, 2024, pp. 1--6.

\bibitem{10283517}
B.~Saglam \emph{et~al.}, ``{Deep Reinforcement Learning Based Joint Downlink Beamforming and RIS Configuration in RIS-Aided MU-MISO Systems Under Hardware Impairments and Imperfect CSI},'' in \emph{2023 IEEE International Conference on Communications Workshops (ICC Workshops)}, 2023, pp. 66--72.

\end{thebibliography}

\end{document}